# Is it personal? The impact of personally relevant robotic failures (PeRFs) on humans' trust, likeability, and willingness to use the robot

Romi Gideoni*, Shanee Honig, Tal Oron-Gilad (ORCID 0000-0002-9523-0161)

Ben-Gurion University of the Negev, Industrial Engineering and Management, Be'er-Sheva, Israel.

*corresponding author: Romigid@post.bgu.ac.il

## Abstract

In three laboratory experiments, we examine the impact of personally relevant failures (PeRFs) on users' perceptions of a collaborative robot. PeR is determined by how much a specific issue applies to a particular person, i.e., it affects one's own goals and values. We hypothesized that PeRFs would reduce trust in the robot and the robot's Likeability and Willingness to Use (LWtU) more than failures that are not personal to participants. To achieve PeR in human-robot interaction, we utilized three different manipulation mechanisms: A) damage to property, B) financial loss, and C) first-person versus third-person failure scenarios. In total, 132 participants engaged with a robot in person during a collaborative task of laundry sorting. All three experiments took place in the same experimental environment, carefully designed to simulate a realistic laundry sorting scenario. Results indicate that the impact of PeRFs on perceptions of the robot varied across the studies. In experiments A and B, the encounters with PeRFs reduced trust significantly relative to a no failure session. But not entirely for LWtU. In experiment C, the PeR manipulation had no impact. The work highlights challenges and adjustments needed for studying robotic failures in laboratory settings. We show that PeR manipulations affect how users perceive a failing robot. The results bring about new questions regarding failure types and their perceived severity on users' perception of the robot. Putting PeR aside, we observed differences in the way users perceive interaction failures compared (experiment C) to how they perceive technical ones (A and B).

## Statements and Declarations

No research funding was received for conducting this study. Students were supported by scholarships as noted in the acknowledgments.

## Acknowledgments

We acknowledge Hahn-Robotics for their donation of the Sawyer robot. The Ben-Gurion University of the Negev provided partial support of the first two authors through the Helmsley Charitable Trust, the Agricultural, Biological and Cognitive Robotics Initiative, the Marcus Endowment Fund, and the George Shrut Chair in Human Performance Management.

# Is it personal? The impact of personally relevant robotic failures (PeRFs) on humans' trust, likeability, and willingness to use the robot

1. Introduction

Robots' technological advancements grow daily, leading to more robots in everyday, ordinary uses [1]. Consumers of personal robots (e.g., robotic vacuum cleaners) actively choose to own or use a robot for a particular task, making it essential to understand what influences people's willingness to interact with 'ordinary use' assistive robots.

Failures influence willingness to interact with a robot [2]. Despite ongoing technological advancements, failures are difficult to eliminate [3],[4]. Human-robot interaction (HRI) studies note types of robotic failures and varying severity levels as impactful [5]–[7]. Yet, most examinations were conducted in artificial or restricted laboratory settings [8], evaluating failures that are not necessarily consequential to participants [2]. In these cases, participants may not have real-life connections to the situation, and therefore their evaluation of failure or its severity is not representitve. Failure situations can be affected by users' motivation and motives, but knowledge of their impact is missing. Personal relevance (PeR), defined as the "level of involvement with an object, situation or action" [9, p. 211], has been found previously to impact user preferences and perceptions of robots (e.g., [10]). Yet, we are unaware of studies on the role of PeR in how robotic failures are perceived and experienced. The literature confirms that PeR affects human perceptions [9] which is why we speculate that it will affect robotic failures' perceptions. Further, PeR as a mitigating factor may also clarify the impact of other factors, such as failure severity.

This research examines the influence of personal relevance (PeR) on inexperienced users' perceptions of a non-perfect laundry sorting assistive robot, while varying the level and type of PeR during a situation of failure. Specifically, we aim to learn whether manipulations of PeR influence the level of trust in the robot (**RQ1**) and likeability and willingness to collaborate with it (**RQ2**). We present three in-person HRI experiments. Utilizing the same experimental setup of a non-perfect laundry sorting assistive robot, each one of the experiments used a different method of achieving PeR.

The first experiment looks at the risk of damage to personal belongings. We created PeR by asking participants to bring clothing items from home. We let the robot perform the laundry sorting task on clothing items of the lab and clothes that participants brought to the experiment. The risk of damage to personal belongings as a PeR manipulation is rooted in the literature. Felt involvement with an item impacts cognitive and psychological arousal, yielding a more significant proportion of product-related thoughts [9]. Products vary from high PeR to low PeR, yet one's belongings tend to be of higher importance, e.g., Cunningham et al. [11], who showed that the level of PeR of objects was related to memory. Participants sorted picture cards into two different baskets: one for their belonging and the second for belonging to someone else. Participants recalled pictures in the self-owned basket better. Further, self-owned possessions are part of the definition of self [12]. The symbolic self-completion theory [13] suggests that the definition of self through material possessions also communicates this identity to others. Therefore, the risk of damaging personal belongings should achieve the negative impact required in this research.

In the second experiment, following Tversky and Kahneman's [8] dependence model for loss aversion, participants are given a sum of money when they enter the lab and experience financial loss if they or the robot linger in completing the laundry sorting task. Robotic failures in such situations, e.g., the robot getting stuck or moving slower than usual, affect the task's duration and contribute to financial loss. In light of Prospect Theory (PT), decision-making is based more on the feeling of loss than the objective outcome [14]. Hence, if the feeling of ownership impacts PeR, it is valid for money as well. The combination of creating the sense of loss and affecting participants' finances requires creating an ownership feeling. We achieve this using the pre-paid mechanisms. Rosenboim and Shavit [15] showed differences in participants' behaviors, most notably a more significant effort to reduce losses by participants who received a prior incentive (i.e., the pre-paid mechanism).

In the third experiment, we compare interaction/communication failures of the robot (e.g., addressing a female as a male), looking at whether the robot erred toward the first person (the participant) or a third-person (an experimenter). We achieve PeR through the different views of actors (first person) and observers (third persons). Jones and Nisbett [16] suggested that actors explain situations through situational factors, while observers' outlook explains the same conditions by attributing them to the actor's characteristics. Many behavioral studies confirmed this notion. Hung and Mukhopadhyay [17] examined this idea in terms of emotional experiences. They showed differences in emotional appraisals given by actors and observers. Regardless of the nature of emotions induced in the reaction (positive or negative), the intensity of the response was more remarkable for the first person. More apparent and extreme emotions suggest that first-person experiences are more personally relevant and less distant than third-person experiences.

The remainder of the paper is structured as follows. Section 2 presents the theoretical grounding for our research. Our general methodology for the three experiments is in section 3. Sections 4-6 present experiments 1 through 3, respectively. Section 7 gives a statistical overview of the outcomes of the three experiments. Section 8 includes the general discussion, limitations, recommendations for future studies, and conclusions.

## 2. Literature review

### 2.1. Personal Relevance

Personal relevance is a well-known concept in psychology. PeR is determined by how much a specific issue is relevant to a particular person, i.e., how the issue affects one's own goals and values [9]. Studies have shown that high PeR motivates information processing [18]. Petty and Cacioppo [18] presented students with arguments from a Princeton professor and a high school teacher on the possibility of introducing changes to their course's curriculum. After hearing the arguments, students listed as many of the opinions provided in the communication as they could remember. When told that the changes would not be happening in the upcoming ten years (i.e., irrelevant for them, low PeR), students were leaning towards the professor's arguments, disregarding the content of his arguments. However, when told changes would happen in the upcoming year (i.e., high PeR), they paid more attention to the content, regardless of the speaker. Hence, high PeR is a motivation factor in processing information [18].

Neurologists have also studied the impact of PeR. Bayer's et al. [19] examined how it affected people's emotional processing by measuring EEG. Participants read sentences in two versions: with high PeR (referencing someone close to the participant) and low PeR (referencing a stranger). Each sentence had either negative, positive, or neutral meanings. Results indicated that people's attention and perceived arousal were greater when reading sentences with high PeR. Sentences with high PeR and non-neutral meanings, led to increased signs of arousal. Hence, familiarity (PeR) and emotional significance cause higher awareness and boost attention [19].

### 2.2. Failures

We adopt Brooks' operational definition of failure as "*a degraded state of ability which causes the behavior or service being performed by the system to deviate from the ideal, normal, or correct functionality*" [16, p. 9]. Since we aim to understand human perceptions, we examine robotic failures that users seek to overcome, since it interferes with their goals. Honig and Oron-Gilad [2] taxonomy of robotic failures distinguishes technical failures from interaction failures. Technical failures are caused by hardware errors or problems in the software. In contrast, interaction failures arise from uncertainties in the robot's interaction with the environment, other agents, and humans [2]. In the human-robot interaction (HRI) domain, studies have examined types of failures and how they affect human users [6, 21-28]. Most investigated failures are technical failures, and only a few studies looked at interaction-related failures [2]. Technical and interaction failures hamper HRI in various ways.

*The impact of failures on trust*

Trust affects people's reliance on automated systems and robots [8]. Trust, is defined as "an attitude which includes the belief that the collaborator will perform as expected, and can, within the limits of the designer's intentions, be relied on to achieve the design goals" [8, p. 3]. In the field of HRI and Human-System Interaction [29], studies show a tendency for a negative influence of failures on people's trust in the robot [6, 8, 25, 30]. Flook et al. [8] evaluated two kinds of failures during a joint assembly task with the participant: *technical failure* (the robot knocking items off on the table) and *decision-level failure* (incorrect assembly guidance; an interaction failure). They found no difference in the perceived level of trust between the two failures types, suggesting that users distinguish between failures by criteria other than failure type. Therefore, researchers examined the influence of failures with different severity levels. Rossi et al. [30] found a positive correlation between the severity of the error performed by the robot and the degree of humans' distrust in it. Yet, the study used a hypothetical interactive storyboard rather than real-world interactions. Another study that did hold real-world interactions [25], suggests that even simple failures negatively impact trust. On the contrary, Van Waveren et al. [6]'s evaluation showed that failures hurt trust compared to flawless interaction. Yet, the negative correlation between trust and the severity of the failure was insignificant. The authors point out that Rossi's failures had a lasting effect (e.g., the robot puts the phone in the toaster), while the failure in their study carried a short-term impact. With no lasting effect, it has no major consequences and is therefore low in PeR. Hence, failures' severity affect humans' trust, yet it is not necessarily the sole factor for the adverse impact.

*The impact of failures on willingness to use (WtU) the robot*

Short-term willingness to use refers to collaboration with the robot right after it has failed. Long-term WtU, relates to future contact intentions towards the robot. Salem et al. [25] researched the impact of failures on WtU of a home-assistance robot. They found that the faulty condition did not impact participants' willingness to help the robot with the future task [25]. Bajones et al. [26] reported that users are willing to help an assistive robot fulfill a task when a failure occurs [26]. These studies provide positivity.

Salem et al. [31] found that people's WtU increases after interacting with a faulty robot. Participants in the faulty incongruent condition rated their willingness to live with the robot and their wish to use it higher than that of the flawless robot [31]. Rosenthal et al. [32] evaluated short- and long-term effects, showing that humans' willingness to help the robot increased when it asked for help.

None of the robot failures presented and examined in the aforementioned studies impacted the user (low PeR), and little is known about how PeR and robot failures affect WtU. A recent study [33] comparing how positive and negative emotions affect willingness to interact with robots suggests that positive emotions like excitement and sympathy are more influential than negative ones like anxiety [33]. When readdressing the studies on failures, the excitement of helping the robot or the empathy it attains by failing may be more influential than the anxiety its failure caused. While this may explain the previous results presented, it enhances the question of whether the tendency would be similar when robot failures are PeR to the users.

*The impact of failures on likeability*

The common conclusion from the literature is that faulty robots are perceived as more friendly and likable than non-faulty ones [23, 26, 31]. This finding is often explained by a psychological concept called the Pratfall Effect, which states that a person's level of attractiveness increases when making a mistake [23]. Accordingly, a non-perfect robot is perceived as more human-like and less machine-like, making it likable [31]. While these studies and others evaluated differences in robots' likeability regarding faulty behavior, none included erroneous behaviors that substantially influenced participants (i.e., all failures had low PeR).

*Initial indications of Personal Relevance (PeR) in HRI studies*

One study that attempted to investigate failures with high PeR is Morales et al. [27]. In this study, the robot's task was to pack grocery items in a simulated grocery store. Two simulated failures with high PeR occurred during the mission: *property risk* (e.g., the robot pushing the bag and its contents off the table) and *personal risk* (e.g., the robot picking a foam potato and throwing it in the participant's direction). The researchers found that participants were willing to help the robot despite its failures, though their willingness to do so was higher pre-failure occurrence. Moreover, the higher risk caused lower ratings of trust. Morales et al. [27] stated that although their results led the participants to believe there was a risk during the interaction, their attempt to construct a feeling of PeR may not have been achieved due to two experimental limitations. The harmed goods weren't the participants' property, reducing the personal impact of property risk. The slow speed of the robotic arm may have decreased the level of personal risk that participants felt during a failure. It was also apparent that the items thrown at participants did not cause distress since they were foam. Perhaps if the objects appeared more natural, the level of perceived personal risk would have been higher.

We aim to understand how personal relevance impacts people's perception of a robot following a PeRF, as measured in trust, WtU the robot in the future, and its likeability. To the best of our knowledge, apart from Morales et al.'s [27], no study has evaluated PeRFs in HRI. Moreover, the settings influence achieving PeR. We emphasized creating a task and an environment to which participants can relate. Our experimental task was sorting laundry, and accordingly, we built the use case and the background environment. Based on the literature review and the PeR manipulations, we hypothesize the following:

*H1: As a failure increases in personal relevance, the perceived level of trust in the robot will decrease.*

Previous research showed that robots' failures decrease people's trust in robots [8]. PeR refers to how relevant the issue is to a person. As such, we would assume that the decrease in trust found in previous studies would be of a higher rate in cases of PeRFs.

*H2: As a failure increases in personal relevance, the robot's likeability and willingness to use the robot again in the future will decrease.*

Studies showed that a faulty robot is perceived as more likable [31]. Many researchers explain that erring robots are perceived positively by the belief that they are more human-like. However, most studies did not examine high PeR failures. We hypothesize that when a failure is of high PeR, the robot will be perceived more negatively due to the meaningful effects of the PeRF.

### 3. General methodology

#### 3.1. Overview

In three experiments, we evaluated three different manipulations of PeR. We used the same experimental environment and task, keeping the general procedures similar, with a few variations to accommodate the failures and manipulations. Experiment-specific methods appear in sections 4-6. Table 1 shows the characteristics of the experiments.

*Table 1 - Differences and similarities between the three experimental methods.*

|  | **First Experiment** | **Second Experiment** | **Third Experiment** |
|---|---|---|---|
| **Method of achieving PeR** | Damage to property | Financial loss | First-person vs. third person |
| **Failures** | Throwing a clothing item on the floor OR into a trash can (Participant's item OR Lab/experimenter's item) | The robot made five sequential errors: Move slowly AND get stuck AND drop a bottle AND make a weird noise AND take an inefficient route | Wrong gender identification of either the participant OR the experimenter |
| **Failure's type** | Technical failure | Technical failure | Interaction failure |
| **Levels of PeR (Excluding the control group)** | 2 | 3 | 2 |
| **Control group** | Did not experience a planned failure | Experienced all failures, did not experience the money loss consequences | Did not experience a planned failure |
| **Other Independent variables** | Severity level | Failures' order (counterbalanced) | Participant's gender, robot's gender |
| **Background story** | Private laundry sorting robot | Public laundromat | Public laundromat |
| **Experimenter's location during the task** | Inside the experimental environment | Separate room | Separate room |

### 3.2. The Laundry room experimental testbed

The experiments took place in a Dome facility. For the experimental testbed to be as realistic as possible, the environment smelled like a fabric softener. Wardrobe and washing machine images were projected on the dome screen, and bottles of laundry softeners were placed on the sorting table. The laundry sorting robot and setup were in the center of the room. The participant stood in front of the laundry sorting table opposite the robot, farther than the three laundry bins (see Figure 1**Error! Reference source not found.**).

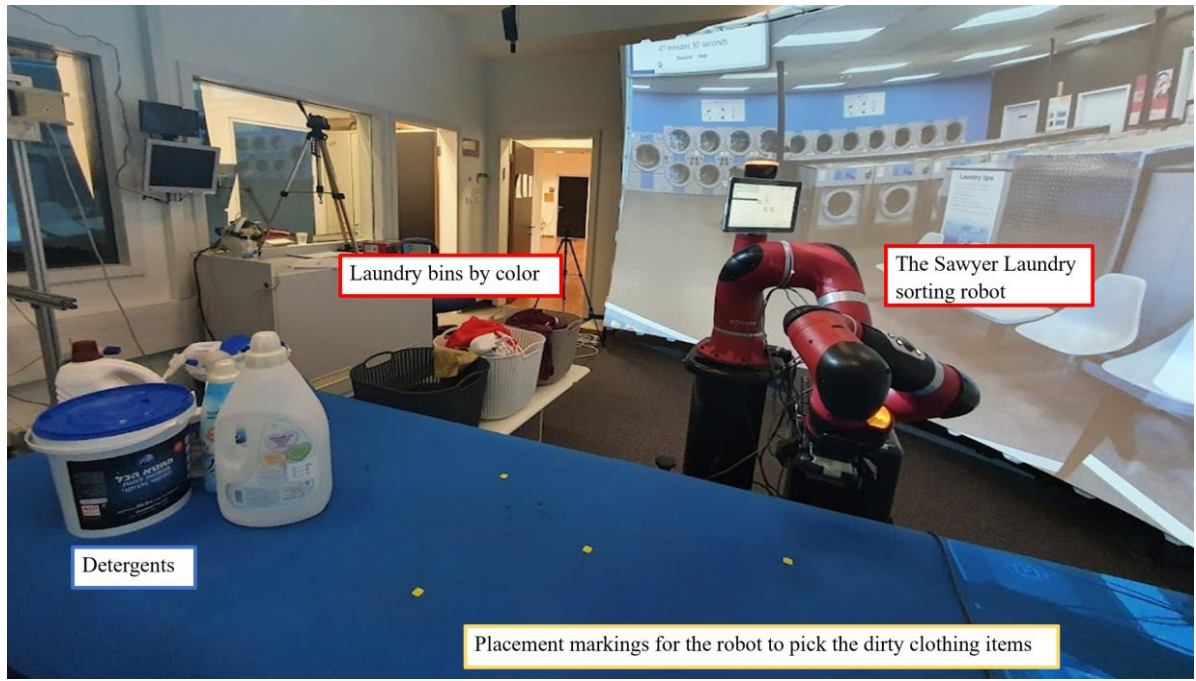

*Figure 1 - The laundry sorting workstation of the robot from the perspective of the participant. Participants stood on the opposite side of the laundry sorting table. To their right was a table with a pile of clothes (not seen in the image).*

### 3.3. The laundry sorting task

Participants stood beside a table with a large pile of "dirty" and "clean" clothes that they had to sort through. All clothes were clean; a blue sticker marked the "dirty" clothes. Clean garments were to be folded and placed on the table in an orderly way. "Dirty" clothes were placed on the robot's laundry sorting workstation for the robot to sort them into the laundry bins. The robot's workstation had five marked collection points where participants could place dirty clothing items (Figure 1). When on a collection point, the robot grabbed an item, examined its color, and put it into three possible laundry bins for white, dark, or colorful items. The laundry sorting task was collaborative in that every time the robot picked an article, the participant placed a new garment instead of it. In addition, participants had to ensure that the robot put each dirty clothing item in the appropriate bin. In case of an erroneous placement, they had to move the clothing item into the correct bin. The task was complete when the "clean" clothing items from the pile were folded, and all dirty items were sorted into the laundry bins.

### 3.4. Apparatus

*The laundry sorting robot*. Sawyer by Rethink Robotics. To enable it to grab clothing items and sort them by color, we replaced its standard gripper with the ROBOTIS HAND RH-P12-RN, which has a wider opening range, and added a Logitech RGB camera above the gripper. Coding in Python was used to integrate the new gripper and camera into the robot's programming with Intera (Sawyer's built-in programming language).

*The Laundry Sorting Algorithm.* Sawyer was programmed to repeatedly and autonomously move along the five designated item collection points on the sorting table. At each point, the robot verified if there was a clothing item, identified its color, picked it up, and moved to place it into the bin that corresponded to the detected color. The robot could identify four different colors: black, white, red, and green. The robot's movements between the points, to and from the laundry bins, and its sorting procedure were autonomous.

*Failure generating procedures.* The experimenter controlled the failures that occurred during the task remotely by initiating a command. This way, we could ensure that the experimenter has control over the order and timing of the failures.

### 3.5. General Procedure

The experimental procedures differed among the three experiments, yet there are many commonalities. Participants arrived one at a time at the lab. At the beginning of the experiment, participants filled the initial questionnaires (InQ), including demographics (age, gender). Then, they were introduced to the robot, tasks, and experimental environment. The experimenter presented the robot as a laundry sorting robot. Participants were told that the robot wasn't working to perfection and may have mistakes. Then, participants received a verbal explanation regarding their collaborative laundry sorting task with the robot.

The laundry sorting tasks occurred in two consecutive sessions: first, with no planned robot failures (no failure session - "NFS"), and then with predesigned robotic failures (failure session - "FS"). Participants performed the same collaborative task with the robot (described in Section 3.2) in the NFS and FS, with a different pile of clothes. After the NFS was complete, participants filled questionnaires. Then, the FS began. During the FS, unknown to the participants, planned failure(s) occurred, depending on the experiment and the participant's experimental condition. Since prior studies showed that the timing of a failure impacts people's perception of the robot ([21, 30, 34]), participants experienced the planned failures at about the same time within the FS. At the end of the FS, participants filled another series of questionnaires. Questionnaires were in Google forms or Qualtrics and filling was conducted on a designated laptop stationed in the lab.

### 3.6. Measures

*Trust.* Since trust is a multifaceted construct, we combined questions from various validated questionnaires in Human-Computer Interaction (HCI); [35], [36]. The questionnaire consisted of five questions, each rated on a Likert scale [1-5], with five the highest. The final score was the sum of ratings after reversing questions 3 and 4. The range is 5-25. To confirm internal consistency between the questions, we calculated Cronbach's alpha using the data from the first experiment and found good reliability ($\alpha=0.821$).

*Future use.* The WtU the robot again was evaluated using one direct statement, which was rated on a Likert scale of [1-5] with five the highest willingness: "I would like to use the robot again."

*Likeability of the robot.* We used a validated questionnaire of likeability in HRI from the GodSpeed questionnaires series [37]. It consisted of five questions, rated on a Likert scale of [1-5]. The final score was the average of ratings, with five the highest.

*Personality questionnaires.* We added a validated questionnaire for each experiment that evaluated participants' particular personal characteristics, which we thought may impact their perceptions of the robotic failure. For each experiment, we evaluated different aspects, depending on the type of failure we examined: Materialism (section 4.3), Risk-taking (section 5.2), Self-efficacy in HRI (section 5.2), and Empathy (section 6.2).

*Stress State Questionnaire (DSSQ; [38]).* We used DSSQ to evaluate participants' subjective experiences of engagement, distress, and worry.

*The number of unexpected mistakes.* This confounding variable represents the number of times the robot made an unplanned mistake (e.g., if the robot mistakenly sorted an item into the wrong laundry bin). From the participant's perspective, such occasions probably count as failures; we wanted to eliminate their impact on our measurements.

### 3.7. Statistical Analyses

Statistical analyses took three stages. Firstly, utilizing paired t-tests, we evaluated whether the measures were statistically different between the NFS and the FS. We did this to ensure that the failure manipulation was successful and impactful enough to produce differentiable results.

Second, we analyzed the impact of PeR by comparing measurements before the failure (BF) and after the failure (AF). Linear regression with backward elimination was used while the dependent variable was the measure AF. In addition, ANOVA tests were conducted to find differences in the primary measures among the PeR experimental conditions. New variables were created to capture numeric differences between the BF measure and the AF measure by subtracting one from the other (i.e., a variable of 'AF minus BF', termed the 'difference variable'). The 'difference variables' were the dependent variables in the ANOVA tests. Post hoc analyses were using Tukey's HSD test. Additional experiment-specific analyses are presented in the descriptions of each experiment (Sections 4-6).

The personal characteristic questionnaires in each experiment resulted in a final score for each participant. We divided participants into two groups using a median split based on their final score (High versus Low). Scores under the median were considered Low, and scores above the median were High.

### 4. Experiment 1 – Risk of damaging personal belongings

### 4.1. Overview

Experiment 1 aimed to evaluate the impact of a failure that may cause damage to participants' belongings and examine the interaction between PeR and perceived failure severity. Failures varied in their level of PeR and severity. The PeRF was of high severity, when the robot dropped a clothing item into the trash can, instead of a laundry bin, or low severity, when the robot dropped a clothing item onto the floor. In the high-severity condition, participants were most likely unaware that the trash can was new, clean, and empty, as it was located farther away from where they stood. The two failure severity levels were chosen based on a survey study [Appendix - A survey study to determine failure severity], that evaluated peoples' perception of the severity level of four separate cases. PeR was manipulated by the clothing item involved in the failure: high PeR was a clothing item belonging to the participant, and low PeR was a clothing item belonging to the lab\experimenter.

### 4.2. Method

*Experimental Design*

A 2X2 between-subject design with the *severity* of the failure (low or high) and *PeR* (low or high) as independent variables. In addition, a control group who did not experience any planned failures in the FS (NF).

*Participants*

Thirty-nine participants (26 Females, 13 Males), 7-8 participants per experimental condition. All participants were Industrial Engineering students at Ben-Gurion University of the Negev who were offered one bonus point course credit as compensation for their participation.

*Experiment-Specific Measures*

1. ***Materialism***. The survey study revealed a correlation between participants' materialistic characteristics and their perception of severity. Therefore, materialistic characteristics were used as a measure via the Three Factors of Materialism questionnaire [39] using the *centrality* factor.
2. ***Attachment to the clothing items***. A 5-point Likert question that evaluated the participants' self-reported attachment to the clothing items they brought.
3. ***Believability***. Since our primary goal was to address personal relevance, we asked participants if the background story given is credible, using a 5-point Likert scale (1- not credible at all, and 5- very believable).
4. ***NARS***. The Negative Attitudes Towards Robots (NARS) is a validated questionnaire that evaluates participants' attitudes towards robots [40].

*Procedure*

Participants were asked in advance to bring five personal clothing items to the experiment: a button dress shirt, a T-shirt, a sweater, and two more pieces of their choice. All items had to be single-colored in black, white, red, or green. When the experiment began, participants were asked to complete the InQ and the NARS. Then, the interactive part started. Participants were given the background story and presented with a pile of 20 clothing items on the table next to them. Participants were asked to imagine that they were in their own home and that this was their pile of clothes, and that their goal was to arrange the pile by sorting it into clean and dirty stacks. Once the sorting was complete, they were asked to add the clothing items they brought to the 'dirty' clothes pile and mix them. Then, the NFS began. It ended when the robot finished sorting eight pieces. Participants then completed the trust questionnaire and DSSQ. The FS that followed varied by the assignment of participants (high or low severity or no-failure).

During the FS, the failure occurred during the robot's pickup of one of the first five clothing items. The PeR failure condition that each participant was assigned to was decided ad-hoc because we couldn't know in advance when a personal clothing item would be placed on a pickup location. If a personal clothing item was placed as one of the first five clothing items the robot collected from the workstation, the participant was assigned to the high-PeR condition. The failure always occurred when the robot picked the personal clothing item. If no personal items were retrieved from the pile during the first five placements, the participant was assigned to the low-PeR condition. In that case, the experimenter initiated a failure when the robot collected the fourth clothing item. The task ended when the 'dirty' stack was sorted. Participants filled questionnaires to measure trust, willingness for future use, and likeability and were debriefed.

### 4.3. Results

*Manipulation check*

Comparisons of the perceived trust BF ($\mu = 19.2, \sigma = 3.12$) and AF ($\mu = 16.6, \sigma = 4.15$) indicated that the manipulation was effective and failures decreased user trust $t(38) = 4.83, p = .00002$.

*Trust*

PeR was a significant predictor of trust ($F(8,30)=4.493, p<.001$), as seen in Table 2. The two-way mixed ANOVA revealed that the interaction between PeR (high/low) and session type (NFS or FS) was statistically significant, $F(2,36)= 4.46, p<.05$, see Figure 3 and Table 3. Both PeR conditions decreased the level of trust in the robot, and participants in the high PeR condition showed greater decrement (*high_PeR: $\mu = -4.75, \sigma = 3.87$; low_PeR: $\mu =-3.6, \sigma = 4.25$*). Participants in the NF condition showed a small increment in trust ($\mu=0.25, \sigma=3.15$) (see Figure 2). Post hoc comparisons revealed significant differences between high PeR and NF and between low PeR and NF. No statistically significant difference was found between high PeR and low PeR.

*Table 2 - Final Trust model in experiment 1 (damage to property)*

|  | **Measure** | $\hat{\beta}$ | $p.v$ |
|---|---|---|---|
|  | Trust BF | 0.318 | 0.097 |
| **Levels of PeR** | PeR 0 | 10.921 | 0.029 |
|  | PeR 1 | -1.889 | 0.282 |
|  | PeR 2 | -5.029 | 0.004 |
|  | Robot's unexpected mistakes | -1.358 | 0.026 |
| **Manipulation-related characteristic measures** | Materialism (high) | -2.966 | 0.026 |
|  | Attachment to Clothings | 1.072 | 0.113 |

*Table 3 - Results of ANOVA test for the impact of personal relevance, session, and their interaction on trust, in the first experiment.*

| Effect | $df_1$ | $df_2$ | F | P |
|---|---|---|---|---|
| **PeR** | 2 | 36 | 0.127 | 0.722 |
| **Session** | 1 | 36 | 16.954 | 0.0002 |
| **PeR X Session** | 2 | 36 | 4.46 | 0.018 |

A two-way ANOVA revealed that there are significant differences in trust between levels of PeR ($F(2, 33) = 4.754, p< .05$) and Materialism ($F(1, 33) = 3.813, p <.05$). Post-hoc analysis revealed that the higher level of Materialism led to a greater decline in trust than a lower level (*diff = -2.658, p=.05*). Further, trust was significantly different between high PeR and NF (*diff= -4.191, p=.01*) and low PeR and NF (*diff= -3.243, p=.06*). There was no significant difference between the two levels of PeR (*p=.67*), see Table 4.

*Table 4 - Top - Results of ANOVA test for the impact of personal relevance and level of Materialism on the difference of trust, in the first experiment. Bottom - Tukey's post hoc results.*

| Effect | $df_1$ | $df_2$ | F | P |
|---|---|---|---|---|
| **PeR** | 2 | 33 | 4.754 | 0.015 |
| **Materialism** | 1 | 33 | 3.813 | 0.05 |
| | Difference of level | Difference of means | $P_{adj}$ | |
| **PeR** | 1-0 | -3.243 | 0.06 | |
| | 2-0 | -4.191 | 0.011 | |
| | 2-1 | -0.947 | 0.67 | |
| **Materialism** | 2-1 | -2.658 | 0.05 | |

*Likeability & Willingness to Use*

We found a strong correlation between likeability and WtU (*Pearson correlation = 0.73*). Therefore, we combined them into one new measure named LWtU (likeability & willingness to use), then used for the subsequent experiments. Since likeability & willingness to use were evaluated only once (AF) and were not normally distributed, we divided the participants into four groups based on their average LWtU scores. This ordinal LWtU measurement was analyzed using ordinal regression. PeR was a significant factor in the final model. Utilizing the Wilcox test, no significant difference between the levels of PeR were found.

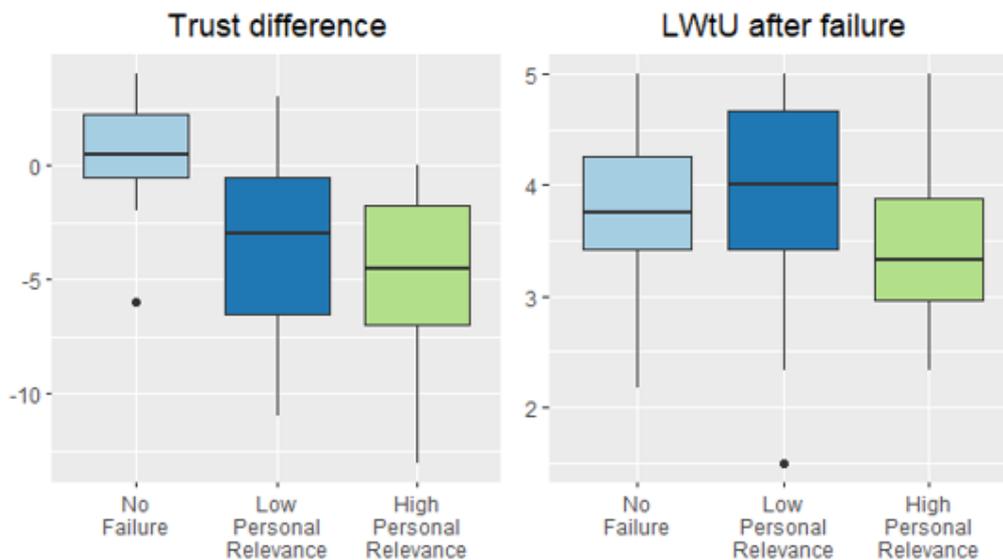

*Figure 2 - Box plots of personal relevance for the risk of damaging a personal object: Left- trust differences (before and after failure), Right - Likability and willingness to use the robot after failure.*

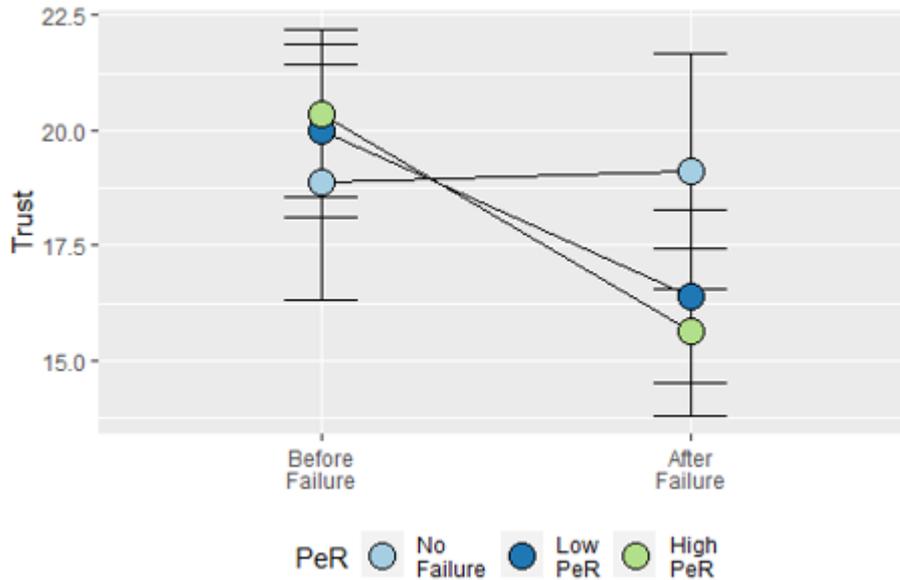

*Figure 3 - The interaction between the level of personal relevance and the session (NFS/FS), in terms of trust, in the first experiment.*

### 4.4. Summary of Experiment 1

We hypothesized that a higher level of PeR would lead to a decrease in trust relative to the low PeR condition. Statistically significant differences were found between each of the PeR conditions and the NF condition. In both PeR conditions, the decrement in trust was greater than in the NF condition. This finding is consistent with prior evidence that robot failures decrease trust [8]. Contrary to what we hypothesized, we did not find statistically significant differences in trust reduction between high and low PeR (Figure 2 and Figure 3).

In the literature, failures tended to increase robot's likeability ([23], [26], [31]). We hypothesized that PeRFs would decrease likeability. The results confirm our hypothesis. When the failure was low in PeR, LWtU was higher than the NF condition, but when PeR was high, LWtU was lower (Figure 2). To further explore this relationship in Experiments 2 and 3, we collected LWtU twice, BF, and AF.

The robot's size and sophistication were not perceived as a realistically affordable household item. Only half of the participants rated the situation credible (20 out of 39 rated Believability as four or higher). For the interaction scenario to be more believable, the robot was presented as a public laundromat robot in Experiments 2 and 3.

## 5. Experiment 2 – Financial Loss due to delays caused by robot failures

### 5.1. Overview

The goal of this experiment was to evaluate the impact of robot failures that cause financial loss. Participants were given a sum of money apriori when entering the experimental session. They were told to use this money to pay for the robot's services during the experiment. Any money leftover would be theirs to keep. The amount of money participants had to pay depends on the amount of time they took to complete the laundry sorting task with the robot. PeR level was determined by the sum of money participants had to pay at the end of the task: assuming that a higher fee enhances the personal relevance. Four PeR conditions were compared, as explained in Table 6.

### 5.2. Method

*Experimental design*

A between-subject design. The amount of money that participants had to pay back at the end of the experiment was manipulated among experimental conditions (Table 6). During the task, each participant experienced five different robotic failures (Table 5). The order of failures was counterbalanced.

*Participants*

Forty participants (16 Females, 24 Males), ages 23-35 ($\mu=25.15$, $\sigma=2.21$), 10 participants per condition. All students at the Ben-Gurion University of the Negev. They received monetary compensation for their participation (at least 50 NIS, but depending on the experimental condition and their responses, they could earn up to 90 NIS more).

*Table 5 – Planned failures that occurred to all participants in the second experiment and their perceived time consequences.*

| Failure | Short | Perceived Consequences |
|---|---|---|
| The robot made a weird and unusual **noise** | Noise | None |
| All of a sudden, the robot started to move very **slowly** | Slow | Between 30-80 additional seconds (depending on when the participants pressed the Reset Button) |
| The robot got stuck and **stopped** moving | Stop | Additional 40 seconds from the moment they pressed the Reset Button |
| The robot dropped an empty **bottle** of fabric softener from the table towards the direction of the participants' feet | Bottle | None |
| The robot took an inefficient longer **route** to the picking point instead of a direct one | Route | Additional 15 seconds |

*Apparatus*

**Reset Button** - Participants could press a "reset button" since we wanted to give them a sense of agency. They were told that if they felt that something was wrong with the robot's behavior, they should press the button to reboot it. When participants pressed the button, the robot returned to the first marked point on the table and continued the sorting process from there. Resetting was done using the Wizard of Oz technique. The experimenter carried out the commands for sending the robot to the first marked point on the table if and when a participant pressed the Reset Button.

**Timer**- During the session, a timer was presented on the dome screen in front of the participants to keep track of time and create a sense of urgency.

**The pricing sheet (PSheet)**- The PSheet reflected three increments in the participant's payment for the robot services (i.e., loss or cost) based on the amount of time the robot was used. The proper PSheet was hung on the wall before the arrival of each participant, depending on the level of PeR they were assigned to. The PSheet included a "first time is free" option so that the NFS would be without any pressure or potential financial loss. The amount of time the robot was used was broken into time slots (TSLs), and the pricing scheme is specified in Table 6.

*PSheet Manipulation and PeR Conditions*

PeR conditions differed from one another by the cost of the robot's service displayed on the PSheet. Practically, participants' performance could not influence the amount of money they had to pay since the robot was the bottleneck of the sorting process. Regardless of the participant's performance, the task's duration depended mainly on the number of clothing items the robot had to sort, and all participants finished in the third TSL. We ensured the duration range by having all participants engage with the same pre-determined number of clothing items.

In the FS session, the PSheet included four TSLs (see Table 6). The first TSL was always free of charge. The cost of using the robot in the following TSLs differed among the four PeR conditions. Duration of use exceeding 12 minutes (TSL3) in PeR1 cost 50 NIS, 70 NIS in PeR2, and 90 NIS in PeR3. In the control condition (PeR0), they did not have to pay, although the duration was similar. This condition is crucial for understanding the impact of financial loss. Participants in the control condition were told that unexpected issues with the robot had caused them to miss out on the previous TSL. Therefore, payment wasn't required. The PSheet of the control group used three TSLs instead of four for simplicity (see Table 6).

Table 6 - The pricing sheet in the second experiment for the personal relevance conditions. Bold and colored is the column that presents the final amount participants had to pay per their assigned PeR. All prices are in New Israeli Shekel (NIS). Recall that participants received 140 NIS at the beginning of the session to cover for the service costs.

| Level of PeR | No Failure Session | Failure Session Cost in NIS per duration in minutes | | | |
|---|---|---|---|---|---|
| | "first use for free" | 0-6 min | 6-12 min | **12-15 min** | Over 15 min |
| 1 | | | 30 | **50** | 60 |
| 2 | 0 | 0 | 50 | **70** | 80 |
| 3 | | | 70 | **90** | 100 |
| 0 (control) | | 0-10 min | | **10-15 min** | Over 15 min |
| | | 0 | | **50 *** | 80 |

*Participants in the control condition were not required to pay.*

*Experiment-specific measures*

1. **Risk-taking (Risk).** To explain the impact of money loss on participants, we evaluate their risk-taking tendencies under the assumption that money loss will have a more negligible effect on high-risk-takers than low-risk-takers. We included the Risk measurement using the GRiPS questionnaire [41].
2. **Self-efficacy in HRI (SE-HRI).** We aimed to make it seem like the robot was responsible for the participants' money loss. Therefore, we wanted to examine if the failures changed participants' self-belief in operating the robot. We assumed that if participants find the robot responsible for their financial loss, their belief in their abilities won't change. To evaluate this, we added a questionnaire of self-efficacy beliefs in interacting with a robot [42]. Self-efficacy was assessed twice, BF and AF.
3. **Feeling of losing money (FinancialLoss).** We added a 5-point scale question to evaluate participants' subjective feelings towards financial loss right after they were required to pay for the robot's services.

In addition, we measured the following objective measurements:

4. The number of times participants pressed the Reset Button.
5. The number of times participants looked at the Timer (as obtained from the video recording of the session).

*Procedure*

First, participants filled the InQ, and the GRiPS questionnaire [41]. Afterward, they were introduced to the experimental environment and told to imagine coming to a public laundromat to do their laundry. They were given 140 NIS (~43 USD) and explained that they would have to use this money to pay for the robot's services during the experiment. Then they were presented with the PSheet of the robot's service according to their PeR condition. Upon understanding the task and the PSheet, the task began. Once completing the NFS, participants filled the SE-HRI questionnaire [42]. Next, they started the FS with the five different failures (Table 5). Failures occurred in equal intervals and controlled by the experimenter. Three failures (*sound*, *slow,* and *stop*) continued until the participant pressed the Reset Button. The other two (*bottle*, *route*) were intermittent, so the robot resumed its regular functioning after the failure occurred. At the end of the FS, the experimenter informed the participant of the service cost, and the participant had to pay back accordingly. Lastly, participants answered the same questionnaires they answered following the NFS. During the debriefing, participants were asked orally regarding their feeling of loss and source of blame.

### 5.3. Results

*Manipulation check*

Comparisons of the perceived trust BF ($\mu = 20.9$, $\sigma = 3.23$) and trust AF ($\mu = 18.25$, $\sigma = 4.02$) indicated that the failures decreased user trust, $t(39) = 4.37$, $p = .00008$. Comparisons of the perceived LWtU BF ($\mu = 4.11$, $\sigma = 0.73$) and LWtU AF ($\mu = 3.71$, $\sigma = 0.99$), $t(39) = 3.8$, $p = .0004$ indicated that failures influenced LWtU.

*Trust*

*As shown in Table 7, a regression equation was formed ($F(9,30)=5.251$, $p<.0005$). The level of PeR manipulation was a significant predictor of trust AF. Further analysis compared groups of PeR regarding trust and revealed significant differences ($F(3, 35) = 2.614$, $p<.05$). Post-hoc tests revealed a substantial difference between PeR2 and PeR0 (diff = -4.5, $p<.05$), other comparisons were insignificant, as shown in*

Table 8 and Figure 4.

*Table 7 - Final Trust model for experiment 2 (financial lossmanipulation of PeR).*

|  | **Measure** | $\hat{\beta}$ | $p.v$ |
|---|---|---|---|
|  | Trust BF | 0.305 | 0.095 |
| **Levels of PeR** | PeR 0 | 14.246 | 0.000 |
|  | PeR 1 | 3.727 | 0.05 |
|  | PeR 2 | -0.24 | 0.885 |
|  | PeR 3 | 4.296 | 0.029 |
| **Personality measures** | Gender (male) | -2.703 | 0.031 |
|  | SE-HRI (high) | 3.932 | 0.003 |
| **Manipulation-related characteristic measures** | Risk (high) | -2.401 | 0.026 |
|  | FinancalLoss | -1.712 | 0.000 |
| **Behavioral measures** | Reset Button presses | 0.578 | 0.098 |

Table 8 - Top - Results of ANOVA test for the impact of personal relevance, gender, and risk score on the difference of trust, in the second experiment. Bottom – Tukey's post hoc results.

| Effect | $df_1$ | $df_2$ | F | P |
|---|---|---|---|---|
| **PeR** | 3 | 35 | 2.614 | 0.05 |
| **Gender** | 1 | 37 | 0.23 | 0.634 |
| **Risk** | 1 | 37 | 1.49 | 0.23 |
| | Difference of level | Difference of means | $P_{adj}$ | |
| **PeR** | 1-0 | -2.3 | 0.498 | |
| | 2-0 | -4.5 | 0.042 | |
| | 3-0 | -2.2 | 0.536 | |
| | 2-1 | -2.2 | 0.536 | |
| | 3-1 | 0.1 | 0.999 | |
| | 3-2 | 2.3 | 0.498 | |

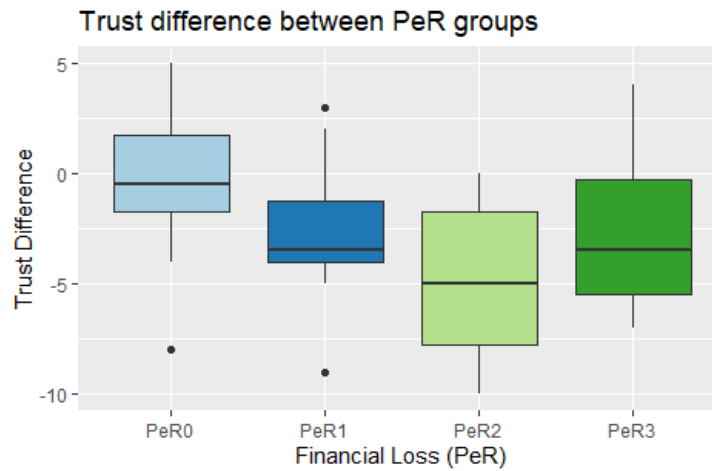

Figure 4 - Trust difference between personal relevance groups in the second experiment

*Likeability & Willingness to Use*

The final model of predicting LWtU AF did not include PeR as an IV. Likewise, a one-way ANOVA found no statistically significant differences between PeR groups in terms of LWtU.

### 5.4. Summary of Experiment 2

Our experimental results align with Experiment 1 that failures with PeR can negatively impact humans' trust in the robot. We hypothesized that a higher PeR (a greater amount of money loss, e.g., PeR 2 or 3), would cause a larger decrease in trust. Our results are inconclusive, as manipulating the highest PeR (PeR3) did not reduce trust more than the other two PeR conditions (Figure 4). That said, in all PeR conditions, the level of trust was lower than in PeR0. Furthermore, aside from the highest PeR group, the trend does seem to align with our hypothesis.

Possibly, participants in the highest PeR group (PeR3) became oblivious to the financial loss because of its high cost. According to their pricing sheet (Table 6), going over the 6 minutes of free-of-charge usage, they were already required to pay a substantial amount of 70 NIS (half of what they received in the beginning), which eventually grew in the following timelapse to 90 NIS. Typically, when a service is time-consuming, prices would increase over the timespan more gradually and reasonably. In the PeR3 instance, usage cost in the first time-lapse was dramatically steeper than the increase in the following time lapses (70 NIS differential followed by 20 NIS and then 10 NIS). In contrast, for PeR1 and PeR2 groups, the raising of the fee was more moderate. This is a paradox – the more severe case became less personally relevant since it was perceived as hopeless and unrealistic. Regarding LWtU, we did not find a statistically significant relationship between PeR and LWtU, which will be further discussed in the general discussion (Section 8).

## 6. Experiment 3 – First Person vs. Third Person

### 6.1. Overview

Here we aimed to evaluate the impact of an interaction failure towards the participant or a third person (one of the experimenters), violating social norms. During the laundry sorting task, the robot spoke to the participant. The conversation was in Hebrew, a gendered language, so addressing another person requires specifying the person's gender. In the high PeR condition, as part of the conversation, the robot incorrectly identified the participant's gender (e.g., approached him as her or vice versa). In the low PeR, the robot incorrectly identified the experimenter's gender (addressing her as he). Eyssel, Bobinger & Ruiter [43] found that people will prefer to interact with a robot that appears as being of their gender [43]. Therefore, half of the participants interacted with the robot when it used a masculine voice, and the other half with the robot using a feminine voice.

### 6.2. Method

*Experimental design*

The experiment used a 2X2X2 between-subject design with the robot's gender (Male or Female), the participant's gender (Male or Female), and the PeR level of the failure (low or high). A high PeRF was the robot saying, *"I am glad **you** are **using** my services. Looks like we are almost done",* while referring to the participant in the wrong gender (The words 'you' and 'using' in Hebrew are gendered). A low PeRF was the robot addressing the female experimenter as a male. Towards the end of the FS, the experimenter left the control room into the Dome environment, supposedly stopping the Go-Pro camera recording (in all conditions). Then, in the low PeR condition, the robot turned to her and said: *"Hello **you**, welcome to the laundromat. Are you next in line? Because we've just finished"*, while referring to a masculine gender. There was also a no failure (NF) control group of participants.

*Participants*

Fifty-three participants (35 Females, 18 Males) participants, ages 22-29 (*μ=25.48, σ=1.64*), 16-19 participants per PeR condition. All were Industrial Engineering students at the Ben-Gurion University of the Negev who were offered one bonus point in a human factors course for their participation.

*Experiment-specific measures*

1. **Empathy.** We wanted to evaluate participants' level of empathy towards others, which may help explain the impact of the failure on their perception of the robot. We used the validated Toronto Empathy Questionnaire (TEQ; [44]).

*Procedure*

Participants filled the InQ at home. When they came to the lab, the experimenter explained the background story (of the public laundromat) and collaborated with the robot to sort clothing items before laundry. The robot then explained how the sorting process happens. During its explanation, it addressed the participants in the correct gender. This was made possible due to the InQ filled previously at home. The questionnaire enabled us to know in advance the participants' gender and set the robot's program accordingly.

After the robot's explanation, the experimenter went into the control room, leaving the participants in the Dome for the NFS. The robot initiated conversation with the participant during the task, using different sentences in each session (see Table 9). Upon completing the NFS, the participant answered the trust, likeability, and future use questionnaires and then moved on to the FS, which lasted around 6 minutes. The interaction error occurred towards the end of the FS for high and low PeR. Finally, upon completion of the FS, the participant was given the last questionnaire, which was identical to the one issued after NFS, with the addition of the TEQ [44].

*Table 9 - The sentences that the robot said during the task. The word 'you' in Hebrew is gendered.*

|  | Sentence |
|---|---|
| **First session (NFS)** | "If **you**'d like **your** clothes to smell nice, **you can** spray the laundry with the Fabric Spray before I pick it up." |
|  | "**You**'re **doing** a great job folding. Well done!" |
| **Second session (FS)** | "This shirt is very dirty. After the laundry, it will be as good as new." |
|  | "I'm glad **you**'re **using** my services. It looks like we're almost done" |
|  | "Hello **you**, **welcome** to the laundromat. Are **you** next in line? Because we've just finished" (only in the low PeR condition) |

### 6.3. Results

*Manipulation check*

We found no difference between trust BF ($\mu = 20.64$, $\sigma = 3.24$) to trust AF ($\mu = 20.41$, $\sigma = 3.33$), $t(52)= 0.525$, $p=0.6$. We found no difference between LWtU BF ($\mu = 4.30$, $\sigma = 0.652$) to LWtU AF ($\mu = 4.35$, $\sigma = 0.651$), $t(52)= -0.876$, $p=0.385$. This indicates that the manipulation of the interaction PeRF had no impact on participants.

### 6.4. Summary

The results show that we chose an unsuitable failure to examine PeRFs. Our initial goal was to select a social-norm-violation type of failure. We can now state that gender misidentification was not a substantial enough issue to make a difference. It became apparent during the experimental runs since, at the end of the FS sessions, we asked participants if they noticed the robot's failure and the change of gender. The vast majority of participants acknowledged that they hadn't even noticed this failure. Further discussion will be given in the general discussion.

## 7. Data across the three studies

We provide an overview of the results across the three studies. Table 10 summarizes the trust-related results, and Table 11 the LWtU.

*Table 10 - An overview of trust measures in all three experiments for all personal relevance conditions.*

|  |  | Trust BF | | Trust AF | | Trust Difference (AF minus BF) | |
|---|---|---|---|---|---|---|---|
|  |  | Mean | sd | Mean | sd | Mean | sd |
| Exp. 1 | PeR0 | 18.88 | 3.72 | 19.13 | 4.09 | 0.57 | 3.26 |
|  | PeR1 | 20.00 | 3.12 | 16.40 | 4.22 | -3.60 | 4.26 |
|  | PeR2 | 20.38 | 2.90 | 15.63 | 3.84 | -4.75 | 3.87 |
|  | **Overall** | **19.92** | **3.12** | **16.64** | **4.15** | **-3.32** | **4.29** |
| Exp. 2 | PeR0 | 19.30 | 1.89 | 18.90 | 4.51 | -0.40 | 3.81 |
|  | PeR1 | 20.60 | 3.63 | 17.90 | 3.38 | -2.70 | 3.47 |
|  | PeR2 | 22.40 | 3.50 | 17.50 | 4.88 | -4.90 | 3.41 |
|  | PeR3 | 21.30 | 3.27 | 18.70 | 3.62 | -2.60 | 3.78 |
|  | **Overall** | **20.90** | **3.23** | **18.25** | **4.02** | **-2.65** | **3.83** |
| Exp. 3 | PeR0 | 19.44 | 2.97 | 20.38 | 3.58 | 1.19 | 3.27 |
|  | PeR1 | 20.11 | 3.69 | 20.53 | 3.24 | -0.17 | 2.83 |
|  | PeR2 | 22.16 | 2.50 | 20.35 | 3.38 | -1.37 | 2.93 |
|  | **Overall** | **20.64** | **3.25** | **20.42** | **3.33** | **-0.19** | **3.13** |

*Table 11 - An overview of likability and willingness to use the robot measure in all three experiments for personal relevance conditions. The first experiment measured likability and willingness to use the robot only once after the failure occurred.*

|  |  | LWtU BF | | LWtU AF | | LWtU Difference (AF minus BF) | |
|---|---|---|---|---|---|---|---|
|  |  | Mean | sd | Mean | sd | Mean | sd |
| Exp. 1 | PeR0 | Not measured | | 3.79 | 0.93 | | |
|  | PeR1 | | | 3.86 | 1.06 | | |
|  | PeR2 | | | 3.45 | 0.82 | | |
|  | **Overall** | | | **3.68** | **0.94** | | |
| Exp. 2 | PeR0 | 3.95 | 0.86 | 3.92 | 1.03 | -0.03 | 0.57 |
|  | PeR1 | 4.30 | 0.60 | 3.57 | 1.25 | -0.73 | 0.81 |
|  | PeR2 | 4.18 | 0.87 | 3.65 | 1.12 | -0.53 | 0.51 |
|  | PeR3 | 4.03 | 0.60 | 3.73 | 0.55 | -0.30 | 0.61 |
|  | **Overall** | **4.12** | **0.73** | **3.72** | **0.99** | **-0.40** | **0.67** |
| Exp. 3 | PeR0 | 4.18 | 0.62 | 4.36 | 0.57 | 0.19 | 0.36 |
|  | PeR1 | 4.17 | 0.63 | 4.27 | 0.59 | 0.10 | 0.21 |
|  | PeR2 | 4.54 | 0.66 | 4.43 | 0.79 | -0.11 | 0.56 |
|  | **Overall** | **4.31** | **0.65** | **4.36** | **0.65** | **0.05** | **0.42** |

## 8. General Discussion

This work aimed to understand how personal relevance impacts people's perception of a collaborating robot after it makes personally relevant failures (PeRFs). We measured trust, willingness to use the robot in the future, and its likeability. We used three different manipulation mechanisms for creating personal relevance: risk of damaging personal possessions, financial loss, and first-person vs. third-person perspective. Not all three manipulations were effective, as we elaborate in this discussion. To increase PeR, we designed a unique experimental environment for a laundry sorting task. Its purpose was to facilitate tasks that participants can relate to in an environment that appears plausible, thus making the value of the task and its outcome more pertinent and understandable. We believed that the ecosystem, situation, and specific failures could be personally relevant to participants.

### 8.1. H1: Trust

Overall, as seen in the literature, robots' failures cause a decrement in trust [6], [25], [30]. This finding was apparent in the first two experiments of this study (Table 10). Experiment 1 showed that those who experienced a PerF felt a decrement in trust of the robot. On the contrary, trust amongst participants who had not experienced a failure showed no decrement but instead increased. This finding supports others in the literature on the increasing trust in robots, as the human interaction with them is longer [30]. In Experiment 2, all participants experienced failures, but only one group wasn't affected by them (Table 6). This group showed only a slight decrement in trust relative to the other groups, who demonstrated a significant decrement in trust. In the first experiment, not all the participants experienced failure. In the second experiment, all participants shared the same failures, yet not all were equally affected. This differentiation is crucial because it allows us to map more accurately the impact of failures on trust. When there is no failure, there is an increment in trust. Failure occurrences lead to a decrement in trust, which becomes more pronounced when the failure is a PeRF.

Contrary to the first two experiments, experiment three examined an interaction failure. The results were different as no correlation between failure and trust was found (Table 10). We think this is due to the failure we chose for the PeRF, misidentifying the participant or experimenter's gender. Although the literature notes differences in personal relevance between first and third person, the manner of portrayal in the failure chosen was non-impactful on participants. We assume that if the robot had revealed more sensitive personal information in a first- or third-person scenario, it would have created a difference between the affected individual and those who were not affected by the error. More personal forms of error may have raised ethical issues. Our findings highlight that interaction failures differ from technical failures. Putting personal relevance aside, we observed differences in the way users perceive interaction failures compared to how they perceive technical ones.

### 8.2. H2: Likability and Willingness to use the robot

The correlation between likeability and the willingness to use the robot (*Pearson correlation = 0.73*), apparent in the first experiment, indicates that these are not necessarily two separate constructs. Therefore we chose to use the LWtU combined measure for our analyses. We hypothesized that contrary to the literature ([23], [31]), a failure high in personal relevance would cause a decrement in likeability. This hypothesis was confirmed in the first experiment (Table 11). For a high severity high

PeRF, the robot's likeability was lower than when there was no failure. Yet, the second experiment did not reveal similar results. No difference in LWtU was found regarding the impact of failures among the PeR manipulation groups (Table 11).

The inconsistency between the results of the two experiments leads us to suggest that to affect likeability negatively, a higher degree of PeRF has to occur than the ones examined in the second experiment. The literature shows that individuals view personal belongings, like clothes in the first experiment, as part of their identity [12]. Accordingly, we can retrospectively consider the difference between the risk of damaging random garments instead of self-owned clothes (i.e., the first experiment) as more significant for personal relevance than the financial loss manipulations in the second experiment (Table 6).

The third experiment revealed near meaningless results regarding the impact of the failure on LWtU (Table 11). While contradicting the hypothesis, these results strengthen our earlier discussion in the trust section that the particular failure we chose was not personally relevant enough to depict the possible impact of first- vs. third-person interaction in the literature.

### 8.3. Differences between findings on Trust and findings on Likability and Willingness to use the robot

Unlike likeability, trust was more affected by the PeR manipulations in experiments 1 and 2. Therefore, the differences between these constructs are discussed more profoundly. The literature indicates that failures tend to increase likeability ([23], [31]) and decrease trust ([6], [25], [30]). It is reasonable to assume that a more impactful failure is required to influence likeability than trust. Therefore, we view the depiction of PeR in the second experiment as impactful enough to decrease trust, yet insufficient to influence likeability and willingness to use the robot again. Further research of PeRFs is needed to learn their impact on likeability and the willingness to use the robot again. We established that affecting likeability and willingness to use the robot requires failures with higher personal relevance than those affecting trust. Due to this, and to the previously discussed indifferent results in the third experiment regarding trust, it makes sense that Experiment 3 also revealed indifference in likeability. Therefore, we still argue that for an interaction failure with a mildly higher depiction of PeR, trust would be affected in a greater fashion than likeability and willingness to use the robot.

### 8.4. Conclusions, Limitations, and Future Work

In conclusion, this study expanded the knowledge on the effects of failures and PeRFs, on human perceptions of robots in face-to-face interactions. There is a difference in the tendency of human perception towards robots when the failure achieves the feeling of PeR. These differences include a seemingly significant effect on human trust, which is a more significant effect than the decrease in trust caused by non-PerFs. While the decrease in trust after the occurrence of a PeRF was more clear-cut, LWtU was not affected in an explicit negative manner as hypothesized. Nonetheless, we conclude that LWtU could indeed be harmed by a PeRF when it is of a high level of PeR. This is different than the effect of a non-PeRF, which generally leads to an increase in LWtU. In addition, we conclude that the difference in results between the various experiments suggests that technical failures may have a more significant impact on human trust and LWtU than interaction failures.

Apart from the limitations of the experimental manipulations (e.g., money gaps in PeR 3 in experiment 2, ethical boundaries in conveying personal information by the robot in experiment 3, etc.), we experienced COVID-19 limitations, mainly regarding the number of participants and the possibility to test a variety of different populations. Since this study focuses on human perceptions, limitations as these can be harmful. The pandemic restrictions significantly affected the third experiment that was initially designed to include a multi-person environment. Instead, the third person was one of the experimenters. For participants, sharing the Dome space with a few unknown people (other participants) rather than being alone in the room may have enhanced the effects of the interaction failure and the psychological significance of the difference between first-person and third-person.

In the future, more studies on PeRF should be performed with time-period adjustments. For instance, this study was designed before the COVID-19 pandemic outburst but was conducted during it. This may have affected the perception of what is essential to people, i.e., what is personally relevant, has been reprioritized. For example, health has become more significant in people's perceptions during this time. Addressing sanitary or health issues in an experiment could have improved our understanding of personally relevant failures during a pandemic. Thus, to further understand the impact of personal relevance on the design of assistive robots, it is crucial to understand, prioritize and conduct studies on what is currently personally relevant to users.

## 10. Appendix - A survey study to determine failure severity

The survey aimed to identify how non-expert users perceive failures as high-severity or low-severity. The outcomes were used to form the experimental failures in Experiment 1. The survey described four different failure outcomes which participants needed to rank according to their perceived severity: (1) *Water:* Water spilling on a clothing item, (2) *Trash:* a clothing item being thrown into the trash can by mistake, (3) *Flowerpot:* a flowerpot falling on their clothing item, (4) *Floor:* a clothing item dropped on the floor. We decided to adopt the highest failure outcome as the high severity condition in Experiment 1 and a lower yet impactful failure outcome as the low-severity case.

*Participants*

Respondents included a convenience sample of 129 people (81 Male, 48 Female), ages 16-67 ($\mu=24.98$, $\sigma=8.53$). Participants were recruited using social media platforms with no inclusion criteria. Participants did not receive compensation for their participation.

*Design*

The survey was created using Google forms and included two parts. The first part collected demographic information (e.g., age, gender) and the respondents' materialistic characteristics via the Three Factors of Materialism questionnaire [41] using the *centrality* factor. We assumed that respondents' rated importance to their possessions as measured by the Materialism questionnaire could be a possible explanatory measure of severity ranking.

The second part of the questionnaire presented short descriptions of the different failure outcomes. Participants rated them in terms of severity. The situations were presented out-of-context, without referencing the robot or the error, to prevent confounding variables. Ratings were on a 5-point Likert scale, where 1=not severe at all and 5=very severe. The four outcomes were presented separately (on different pages) to prevent direct comparisons between them.

*Analysis and Results*

*Trash* (a clothing item being thrown into the trash can by mistake) was perceived as the most severe ($\mu=3.56$, $\sigma=1.02$), followed by *Flowerpot* ($\mu=2.52$, $\sigma=1.00$) and *Floor* ($\mu=1.90$, $\sigma=0.96$). *Water* was perceived as the least severe ($\mu=1.59$, $\sigma=0.71$). Since we wanted the lowest severity failure in our experiment to have some degree of personal relevance, we chose *Floor* (a clothing item dropped on the floor) as the lowest.

We calculated Spearman correlation for each of the cases. We ran a permutation test with 10,000 permutations in each case to ensure our results are not coincidental. Results are shown in Table 12 and Figure 5. Results ranged from the 98.9th percentile (for *Floor*) to the 100th percentile (for *trash*). Hence, 98.9 percent of the other permutations resulted in a lower Spearman correlation in the Floor condition, indicating that our original results are highly correlated. We adjusted P-Value using the Bonferroni correlation (0.05/4= 0.0125). Since all p. values were lower than 0.0125, we concluded that the results were significant ($p<0.01$). We decided to add the materialism questionnaire to Experiment 1, to evaluate and account for the potential impact of Materialism on failure perceptions.

*Table 12 - Spearman correlation of each case and p-value after permutation test.*

| Case | Spearman Correlation | Permutation test p-value |
|---|---|---|
| **Spilling Water** | 0.2519 | 0.004 |
| **Throwing to the Trash** | 0.3757 | 0.0000 |
| **Falling flowerpot** | 0.3082 | 0.0003 |
| **Throwing on the Floor** | 0.2228 | 0.011 |

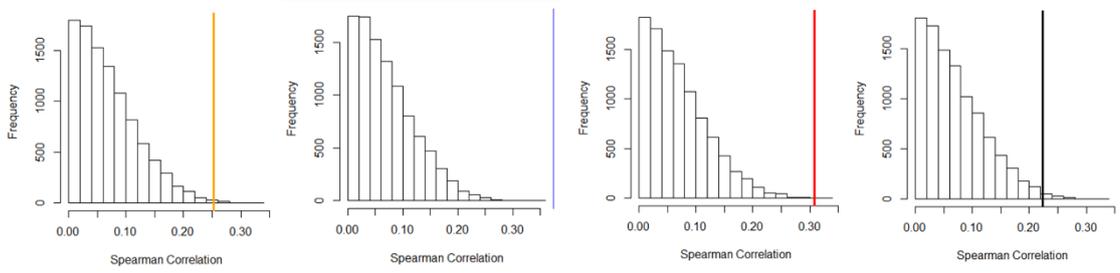

*Figure 5 - results of permutation test. The line represents our spearman correlation. Graphs by cases order: spilling water, throwing to the trash, falling flowerpot and throwing on the floor.*